\documentclass[11pt,a4paper]{article}
\usepackage[hyperref]{acl2017}
\usepackage{times}
\usepackage{latexsym}
\usepackage[font={footnotesize}]{caption}
\usepackage{amsmath}
\usepackage{amssymb}
\usepackage{multirow}
\usepackage{color}
\usepackage{booktabs}
\usepackage{enumitem}
\usepackage{graphicx}
\usepackage{url}

\newcommand\nl[1]{{\it``#1''}}

\aclfinalcopy
\title{Evaluating Semantic Parsing against a Simple Web-based Question Answering Model}

\author{Alon Talmor \\ Tel-Aviv University \\  {\small \tt alontalmor@mail.tau.ac.il} \\ 
         \And  Mor Geva \\ Tel-Aviv University \\  {\small \tt morgeva@mail.tau.ac.il} \And
         Jonathan Berant \\ Tel-Aviv University \\ {\small \tt joberant@cs.tau.ac.il}}
\date{}

\begin{document}

\maketitle

\begin{abstract}
Semantic parsing shines at analyzing complex natural language that involves composition and computation over multiple pieces of evidence. However, datasets for semantic parsing contain many factoid questions that can be answered from a single web document. 
In this paper, 
we propose to evaluate semantic parsing-based question answering models by comparing them to a  question answering baseline that queries the web and extracts the answer only from web snippets, without access to the target knowledge-base. We investigate 
this approach on \textsc{ComplexQuestions}, a dataset designed to focus on compositional language, and find that our model obtains reasonable performance ($\sim$35 F$_1$ compared to 41 F$_1$ of state-of-the-art). We find in our analysis that our model performs well on complex questions involving conjunctions, but struggles on questions that involve relation composition and superlatives.
\end{abstract}

\section{Introduction}
Question answering (QA) has witnessed a surge of interest in recent years \cite{hill2015goldilocks,yang2015wikiqa,pasupat2015compositional,chen2016thorough,joshi2017triviaqa}, as it is one of the prominent tests for natural language understanding. QA can be coarsely divided into semantic parsing-based QA, where a question is translated into a logical form that is executed against a knowledge-base \cite{zelle96geoquery,zettlemoyer05ccg,liang11dcs,kwiatkowski2013scaling,reddy2014large,berant2015agenda}, and unstructured QA, where a question is answered directly from some relevant text \cite{voorhees2000building,hermann2015read,hewlett2016wikireading,kadlec2016text,seo2016bidaf}. 

In semantic parsing, background knowledge has already been compiled into a knowledge-base (KB), and thus the challenge is in interpreting the question, which may contain compositional constructions (\nl{What is the second-highest mountain in Europe?}) or computations (\nl{What is the difference in population between France and Germany?}). In unstructured QA, the model needs to  also interpret the language of a document, and thus most datasets focus on matching the question against the document and extracting the answer from some local context, such as a sentence or a paragraph \cite{onishi2016wdw,rajpurkar2016squad,yang2015wikiqa}.

Since semantic parsing models excel at handling complex linguistic constructions and reasoning over multiple facts, a natural way to examine whether a benchmark indeed requires modeling these properties, is to train an unstructured QA model, and check if it under-performs compared to semantic parsing models.
If questions can be answered by examining local contexts only, then the use of a knowledge-base is perhaps unnecessary. However, to the best of our knowledge, only models that utilize the KB have been evaluated on common semantic parsing benchmarks. 

The goal of this paper is to bridge this evaluation gap. We develop a simple log-linear model, in the spirit of traditional web-based QA systems \cite{kwok2001scaling,brill2002askmsr}, that answers questions by querying the web and extracting the answer from returned web snippets. Thus, our evaluation scheme is suitable for semantic parsing benchmarks in which the knowledge required for answering questions is covered by the web (in contrast with virtual assitants for which the knowledge is specific to an application).

We test this model on \textsc{ComplexQuestions} \cite{bao2016constraint}, a dataset designed to
require more compositionality 
compared to earlier datasets, such as \textsc{WebQuestions} \cite{berant2013freebase} and \textsc{SimpleQuestions} \cite{bordes2015simple}. 
We find that a simple QA model, despite having no access to the target KB, performs reasonably well on this dataset ($\sim$35 F$_1$ compared to the state-of-the-art of 41 F$_1$).
Moreover, for the subset of questions for which the right answer can be found in one of the web snippets, we outperform the semantic parser (51.9 F$_1$ vs. 48.5 F$_1$).
We analyze results for different types of compositionality and find that superlatives and relation composition constructions are challenging for a web-based QA system, while conjunctions and events with multiple arguments are easier.

An important insight is that semantic parsers must overcome the mismatch between natural language and formal language. Consequently, language that can be easily matched against the web may become challenging to express in logical form. For example, the word \nl{wife} is an atomic binary relation in natural language, but expressed with a complex binary $\lambda x. \lambda y. \mbox{Spouse}(x, y) \land \mbox{Gender}(x, \mbox{Female})$ in knowledge-bases. Thus, some of the complexity of understanding natural language is removed when working with a natural language representation.

To conclude, we propose to evaluate the extent to which semantic parsing-based QA benchmarks require compositionality by comparing semantic parsing models to a baseline that extracts the answer from short web snippets. We  obtain reasonable performance on \textsc{ComplexQuestions}, and analyze the types of compositionality that are challenging for a web-based QA model. To ensure reproducibility, we release our dataset, which attaches to each example  from \textsc{ComplexQuestions} the top-100 retrieved web snippets.\footnote{Data can be downloaded from \url{https://worksheets.codalab.org/worksheets/0x91d77db37e0a4bbbaeb37b8972f4784f/}}

\section{Problem Setting and Dataset}
\label{sec:data}
\begin{figure}
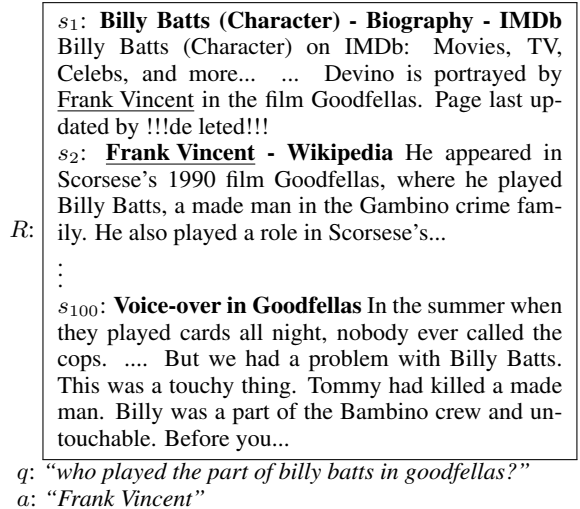

\centering
\noindent
\small 

$R$:
\fbox{

\begin{minipage}{21em}
$s_1$: 
\textbf{Billy Batts (Character) - Biography - IMDb} Billy Batts (Character) on IMDb: Movies, TV, Celebs, and more... ... Devino is portrayed by \underline{Frank Vincent} in the film Goodfellas. Page last updated by !!!de
leted!!!

$s_2$: \textbf{\underline{Frank Vincent} - Wikipedia} He appeared in Scorsese's 1990 film Goodfellas, where he played Billy Batts, a made man in the Gambino crime family. He also played a role in Scorsese's...

\vdots

$s_{100}$: \textbf{Voice-over in Goodfellas} In the summer when they played cards all night, nobody ever called the cops. .... But we had a problem with Billy Batts. This was a touchy thing. Tommy had
killed a made man. Billy was a part of the Bambino crew and untouchable. Before you...
\end{minipage}
}
\begin{minipage}{23em}
\noindent \textbf{$q$}: \nl{who played the part of billy batts in goodfellas?}\\
\noindent \textbf{$a$}: \nl{Frank Vincent}
\end{minipage}
\caption{\small A training example containing a result set $R$, a question $q$ and an answer $a$.
The result set $R$ contains 100 web snippets $s_i$, each including a title (boldface) and text. The answer is underlined.
}
\label{fig:training_ex}
\end{figure}

Given a training set of triples $\{q^{(i)}, R^{(i)}, a^{(i)}\}_{i=1}^{N}$, where $q^{(i)}$ is a question, $R^{(i)}$ is a web result set, and $a^{(i)}$ is the answer, our goal is to learn a model that produces an answer $a$ for a new question-result set pair $(q, R)$. A web result set $R$ consists of $K(=100)$ web snippets, where each snippet $s_i$ has a title and a text fragment. An example for a training example is provided in Figure~\ref{fig:training_ex}.

Semantic parsing-based QA datasets contain question-answer pairs alongside a background KB. To convert such datasets to our setup, we run the question $q$ against Google's search engine and scrape the top-$K$ web snippets. We use only the web snippets and ignore any boxes or other information returned (see Figure~\ref{fig:training_ex} and the full dataset in the supplementary material).

\paragraph{Compositionality}
We argue that if a dataset truly requires a compositional model, then it should be difficult to tackle with methods that only match the question against short web snippets. This is since it is unlikely to integrate all necessary pieces of evidence from the snippets.

We convert \textsc{ComplexQuestions} into the aforementioned format, and manually analyze the types of compositionality that occur on 100 random training examples. Table~\ref{tab:compositionality} provides an example for each of the  question types we found:
ֿ\begin{enumerate}[topsep=0pt,itemsep=0pt,partopsep=0pt,parsep=0pt]
\item[] \textsc{Simple}: an application of a single binary relation on a single entity.
\item[] \textsc{Filter}: a question where the semantic type of the answer is mentioned (\nl{tv shows} in Table~\ref{tab:compositionality}).
\item[] \textsc{N-ary}: A question about a single event that involves more than one entity (\nl{juni} and \nl{spy kids 4} in Table~\ref{tab:compositionality}).
\item[] \textsc{Conjunction}: A question whose answer is the conjunction of more than one binary relation in the question.
\item[] \textsc{Composition} A question that involves composing more than one binary relation over an entity (\nl{grandson} and \nl{father} in Table~\ref{tab:compositionality}).
\item[] \textsc{Superlative} A question that requires sorting or comparing entities based on a numeric property.
\item[] \textsc{Other} Any other question.
\end{enumerate}

Table~\ref{tab:compositionality} illustrates that \textsc{ComplexQuestions} is dominated by \textsc{n-ary} questions that involve an event with multiple entities. In Section~\ref{sec:experiments} we evaluate the performance of a simple QA model for each compositionality type, and find that \textsc{N-ary} questions are handled well by our web-based QA system.

\begin{table}[t]
\begin{center}
\scriptsize{
\begin{tabular}{l|l|c}
 \toprule
 \textbf{Type} & \textbf{Example} & \% \\ 
 \midrule
\textsc{Simple}      & \nl{who has gone out with cornelis de graeff} & 17\%  \\ 
\textsc{Filter}          & \nl{which tv shows has wayne rostad starred in}  &  18\% \\ 
\textsc{N-ary}          &  \nl{who played juni in spy kids 4?} & 51\%  \\ 
\textsc{Conj.}          & \emph{``what has queen latifah starred in that doug}  & 10\%  \\ 
   & \emph{mchenry directed"}  &  \\ 
\textsc{Compos.}          & \nl{who was the grandson of king david's father?} &  7\% \\
\textsc{Superl.}          &  \nl{who is the richest sports woman?} &  9\% \\ 
\textsc{Other}          &  \nl{what is the name george lopez on the show?} &  8\% \\ 
\toprule
\end{tabular}}
\end{center}
\caption{An example for each compositionality type and the proportion of examples in 100 random examples. A question can fall into multiple types, and thus the sum exceeds 100\%.}
\label{tab:compositionality}
\end{table}

\section{Model}
Our model comprises two parts. First, we extract a set of answer candidates, $\mathcal{A}$, from the web result set. Then, we train a log-linear model that outputs a distribution over the candidates in $\mathcal{A}$, and is used at test time to find the most probable answers.

\paragraph{Candidate Extraction}
We extract all 1-grams, 2-grams, 3-grams and 4-grams (lowercased) that appear in $R$, yielding roughly 5,000 candidates per question. We then discard any candidate that fully appears in the question itself, and define $\mathcal{A}$ to be  the top-$K$ candidates based on their tf-idf score, where term frequency is computed on all the snippets in $R$, and inverse document frequency is computed on a large external corpus.

\paragraph{Candidate Ranking}

We define a log-linear model over the candidates in $\mathcal{A}$:
\begin{align*}
p_\theta(a \mid q, R) = \frac{\exp(\phi(q, R, a)^\top \theta)}{\sum_{a' \in \mathcal{A}} \exp(\phi(q, R, a')^\top \theta)},
\end{align*}
where $\theta \in \mathbb{R}^d$ are learned parameters, and $\phi(\cdot) \in \mathbb{R}^d$ is a feature function. We train our model by maximizing the regularized conditional log-likelihood objective $\sum_{i=1}^N \log p_\theta(a^{(i)} \mid q^{(i)}, R^{(i)}) + \lambda \cdot ||\theta||_2^2$. 
At test time, we return the most probable answers based on $p_\theta(a \mid q, R)$ (details in Section~\ref{sec:experiments}). While semantic parsers generally return a set, in \textsc{ComplexQuestions} 87\% of the answers are a singleton set.

\paragraph{Features} A candidate span $a$ often has multiple mentions in the result set $R$. Therefore, our feature function $\phi(\cdot)$ computes the average of the features extracted from each mention.
The main information sources used are the match between the candidate answer itself and the question (top of Table~\ref{tab:features}) and the match between the context of a candidate answer in a specific mention and the question (bottom of Table~\ref{tab:features}), as well as the Google rank in which the mention appeared.

Lexicalized features are useful for our task, but the number of training examples is too small to train a fully lexicalized model. Therefore, we define lexicalized features over the 50 most common non-stop words in \textsc{ComplexQuestions}. Last, our context features are defined in a 6-word window around the candidate answer mention, where the feature value decays exponentially as the distance from the candidate answer mention grows.
Overall, we compute a total of 892 features over the dataset.

\begin{table}[t]
\begin{center}
\scriptsize{
\begin{tabular}{l|l}
 \toprule
 \textbf{Template} & \textbf{Description}  \\ 
 \midrule
\textsc{Span length} & Indicator for the number of tokens in $a_m$    \\ 
\textsc{tf-idf} & Binned and raw tf-idf scores of $a_m$ for every \\ & span length  \\ 
\textsc{Capitalized} & Whether $a_m$ is capitalized  \\ 
\textsc{Stop word} & Fraction of words in $a_m$ that are stop words  \\ 
\textsc{In quest} & Fraction of words in $a_m$ that are in $q$\\
\textsc{In quest+Common} & Conjunction of \textsc{In quest} with common words \\ & in $q$ \\ 
\textsc{In question dist.} & Max./avg.
cosine similarity between $a_m$ \\ & words and $q$ words\\
\textsc{Wh+NE} & Conjunction of wh-word in $q$ and named entity \\ & tags (NE) of $a_m$\\
\textsc{Wh+POS} & Conjunction of wh-word in $q$ and \\ & part-of-speech tags of $a_m$\\
\textsc{NE+NE} & Conjunction of NE tags in $q$ and NE tags in $a_m$\\
\textsc{NE+Common} & Conjunction of NE tags in $a_m$ and common \\ & words in $q$ \\
\textsc{Max-NE} & Whether $a_m$ is a NE with maximal span \\ &  (not contained in another NE) \\
\textsc{year} & Binned indicator for year if $a_m$ is a year \\
\midrule
\textsc{Ctxt match} & Max./avg. over non stop words in $q$, for \\ & whether a $q$ word    occurs around $a_m$, weighted \\ & by distance  from $a_m$ \\
\textsc{Ctxt similarity} & Max./avg. cosine similarity over non-stop \\
& words in $q$, between $q$ words and words around \\
& $a_m$,  weighted by distance \\
\textsc{In title} & Whether $a_m$ is in the title part of the snippet  \\ 
\textsc{Ctxt entity} & Indicator for whether a common word appears \\
& between $a_m$ and a named entity that appears\\
&in $q$\\ 
\textsc{Google rank} & Binned snippet rank of $a_m$ in the result set $R$ \\
\toprule
\end{tabular}}
\end{center}
\caption{Features templates used to extract features from each answer candidate mention $a_m$. Cosine similarity is computed with pre-trained GloVe embeddings \cite{pennington2014glove}. The definition of \emph{common words} and \emph{weighting by distance} is in the body of the paper.}
\label{tab:features}
\end{table}

\section{Experiments}
\label{sec:experiments}

\textsc{ComplexQuestions} contains 1,300 training examples and 800 test examples. We performed 5 random 70/30 splits of the training set for development. We computed POS tags and named entities with Stanford CoreNLP~\cite{manning2014stanford}. We did not employ any co-reference resolution tool in this work. If after candidate extraction, we do not find the gold answer in the top-$K$(=140) candidates, we discard the example, resulting in a training set of 856 examples. 

We compare our model, \textsc{WebQA}, to \textsc{STAGG} \cite{yih2015stagg} and \textsc{CompQ} \cite{bao2016constraint}, which are to the best of our knowledge the highest performing semantic parsing models on both \textsc{ComplexQuestions} and \textsc{WebQuestions}. For these systems, we only report test F$_1$ numbers that are provided in the original papers, as we do not have access to the code or predictions.
We evaluate models by computing average F$_1$, the official evaluation metric defined for \textsc{ComplexQuestions}. This measure computes the F$_1$ between the set of answers returned by the system and the set of gold answers, and averages across questions. To allow \textsc{WebQA} to return a set rather than a single answer, we return the most probable answer $a^*$ as well as any answer $a$ such that $(\phi(q, R, a^*)^\top \theta - \phi(q, R, a)^\top \theta) < 0.5$.
We also compute precision@1 and Mean Reciprocal Rank (MRR) for \textsc{WebQA}, since we have a ranking  over answers.
To compute metrics we lowercase the gold and predicted spans and perform exact string match.

\begin{table}[t]
\begin{center}
\scriptsize{
\begin{tabular}{l|c|c|c|c|c|}
 \toprule
 & \multicolumn{2}{c|}{\textbf{Dev}} & \multicolumn{3}{c|}{\textbf{Test}}\\ 
 \textbf{System} & \textbf{F$_1$} & \textbf{p@1} & \textbf{F$_1$} & \textbf{p@1} & \textbf{MRR} \\ 
 \midrule
\textsc{STAGG} & - & -  & 37.7 & - & -   \\ 
\textsc{CompQ} & - & - & \textbf{40.9} & - & -  \\ 
\midrule
\textsc{WebQA} & 35.3 & 36.4 & 32.6  & 33.5 & 42.4    \\
\textsc{WebQA-extrapol} & - & - & 34.4 & - & -   \\
\midrule
\textsc{CompQ-Subset} & - & - & 48.5 & - & -   \\
\textsc{WebQA-Subset}  & 53.6 & 55.1 & 51.9 & 53.4 & 67.5  \\
\toprule
\end{tabular}}
\end{center}
\caption{Results on development (average over random splits) and test set. Middle: results on all examples. Bottom: results on the subset where candidate extraction succeeded.}
\label{tab:results}
\end{table}

Table~\ref{tab:results} presents the results of our evaluation. \textsc{WebQA} obtained 32.6 F$_1$ (33.5 p@1, 42.4 MRR) compared to 40.9 F$_1$ of \textsc{CompQ}. Our candidate extraction step finds the correct answer in the top-$K$ candidates in 65.9\% of development examples and 62.7\% of test examples. Thus, our test F$_1$ on examples for which candidate extraction succeeded (\textsc{WebQA-Subset}) is 51.9 (53.4 p@1, 67.5 MRR).

We were able to indirectly compare \textsc{WebQASubset} to \textsc{CompQ}: 
\newcite{bao2016constraint} graciously provided us with the predictions of \textsc{CompQ} when it was trained on \textsc{ComplexQuestions}, \textsc{WebQuestions}, and \textsc{SimpleQuestions}. In this setup, \textsc{CompQ} obtained 42.2 F$_1$ on the test set (compared to 40.9 F$_1$, when training on \textsc{ComplexQuestions} only, as we do). Restricting the predictions to the subset for which candidate extraction succeeded, the F$_1$ of \textsc{CompQ-Subset} is 48.5, which is 3.4 F$_1$ points lower than \textsc{WebQA-Subset}, which was trained on less data.

Not using a KB, results in a considerable disadvantage for \textsc{WebQA}.
KB entities have normalized descriptions, and the answers have been annotated according to those descriptions.
We, conversely, find answers on the web and often predict a correct answer, but get penalized due to small string differences. 
E.g., for \nl{what is the longest river in China?} we answer \nl{yangtze river}, while the gold answer is \nl{yangtze}.
To quantify this effect we manually annotated all 258 examples in the first random development set split, and determined whether string matching failed, and we actually returned the gold answer.\footnote{We also publicly release our annotations.} 
This improved performance from 53.6 F$_1$  to 56.6 F$_1$ (on examples that passed candidate extraction). Further normalizing gold and predicted entities, such that \nl{Hillary Clinton} and \nl{Hillary Rodham Clinton} are unified, improved F$_1$ to 57.3 F$_1$. Extrapolating this to the test set would result in an F$_1$ of 34.4 (\textsc{WebQA-extrapol} in Table~\ref{tab:results}) and 34.9, respectively.

Last, to determine the contribution of each feature template, we performed ablation tests and we present the five feature templates that resulted in the largest drop to performance on the development set in Table~\ref{tab:ablation}. 
Note that TF-IDF is by far the most impactful feature, leading to a large drop of 12 points in performance. This shows the importance of using the redundancy of the web for our QA system.

\begin{figure}
\centering
\includegraphics[scale=0.36]{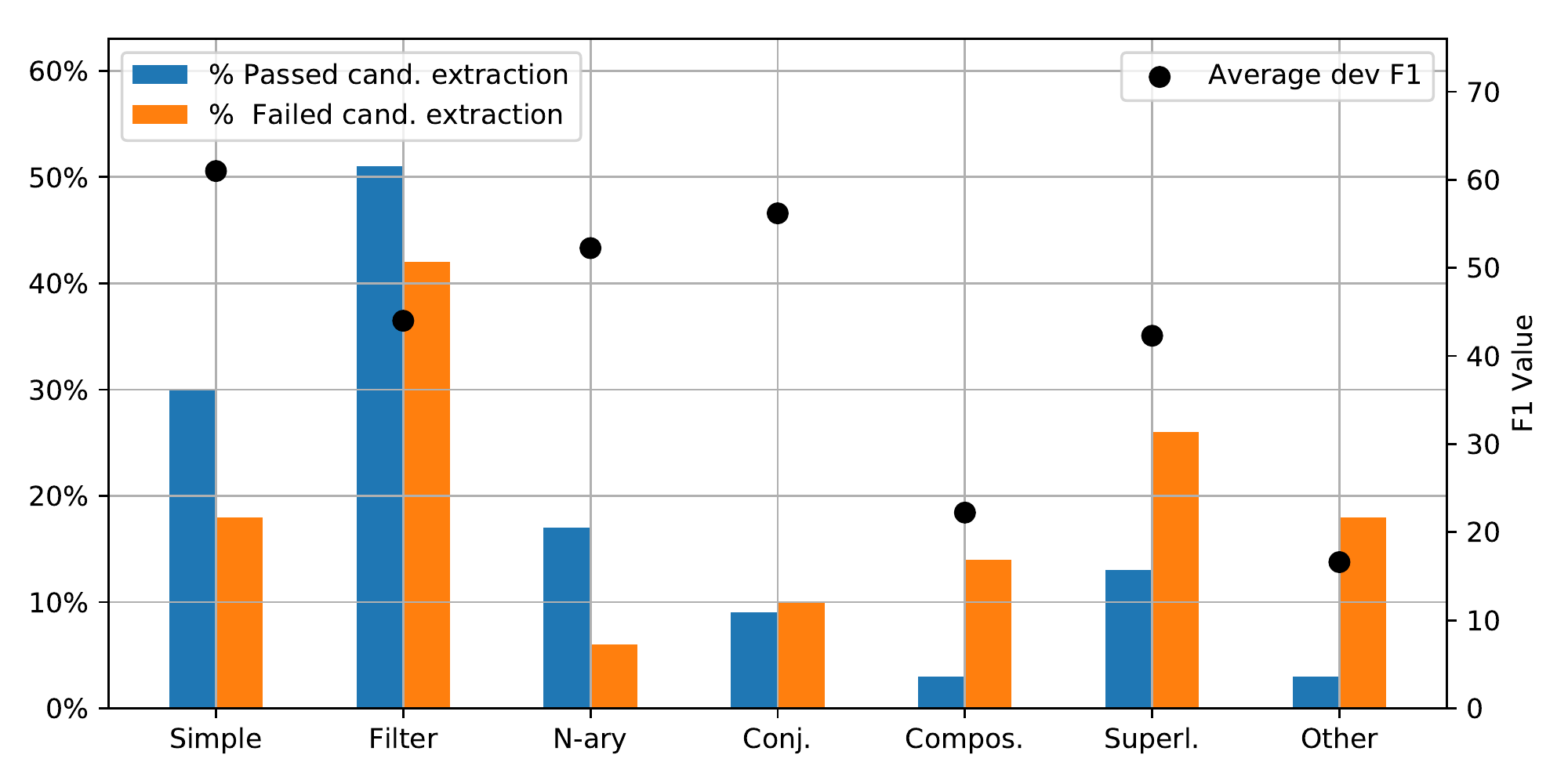}
\caption{Proportion of examples that passed or failed candidate extraction for each compositionality type, as well as average F$_1$ for each compositionality type. \textsc{Composition} and \textsc{Superlative} questions are difficult for \textsc{WebQA}.}
\label{fig:compositionality}
\end{figure}
\begin{table}[t]
\begin{center}
\scriptsize{
\begin{tabular}{l|c|c|c|c|c|}
 \toprule 
 \textbf{Feature Template} & \textbf{F$_1$} & \textbf{$\Delta$} \\ 
 \midrule
\textsc{WebQA} & 53.6 & \\  
\hline
\textsc{- Max-NE} & 51.8 & -1.8 \\
\textsc{- Ne+Common} & 51.8 & -1.8 \\
\textsc{- Google Rank} & 51.4 & -2.2 \\
\textsc{- In Quest} & 50.1 & -3.5\\ 
\textsc{- TF-IDF} & 41.5 & -12\\ 
\toprule
\end{tabular}}
\end{center}
\caption{Feature ablation results. The five features that lead to largest drop in performance are displayed.}
\label{tab:ablation}
\end{table}

\paragraph{Analysis} 
To understand the success of \textsc{WebQA} on different compositionality types, we manually annotated the compositionality type of 100 random examples that passed candidate extraction and 50 random examples that failed candidate extraction. Figure~\ref{fig:compositionality} presents the results of this analysis, as well as the average F$_1$ obtained for each compositionality type on the 100 examples that passed candidate extraction (note that a question can belong to multilpe compositionality types). We observe that \textsc{Composition} and \textsc{Superlative} questions are challenging for \textsc{WebQA}, while \textsc{Simple}, \textsc{Filter}, and \textsc{N-ary} quesitons are easier (recall that a large fraction of the questions in \textsc{ComplexQuestions} are \textsc{N-ary}). Interestingly, \textsc{WebQA} performs well on \textsc{Conjunction} questions (\nl{what film victor garber starred in that rob marshall directed}), possibly because the correct answer can obtain signal from multiple snippets.

An advantage of finding answers to questions from web documents compared to semantic parsing, is that we do not need to learn the ``language of the KB''. For example, the question \nl{who is the governor of California 2010} can be  matched directly to web snippets, while in Freebase \cite{bollacker2008freebase} the word \nl{governor} is expressed by a complex predicate $\lambda x.\exists z. \text{GoverPos}(x,z) \land \text{PosTitle}(z, \text{Governor})$. This could provide a partial explanation for the reasonable performance of \textsc{WebQA}.

\section{Related Work}
Our model \textsc{WebQA} performs QA using web snippets, similar to traditional QA systems like \textsc{Mulder} \cite{kwok2001scaling} and \text{AskMSR} \cite{brill2002askmsr}. However, it it enjoys the advances in commerical search engines of the last decade, and uses a simple log-linear model, which has become standard in Natural Language Processing.

Similar to this work, \newcite{yao2014freebase} analyzed a semantic parsing benchmark with a simple QA system. However, they employed a semantic parser that is limited to applying a single binary relation on a single entity, while we develop a QA system that does not use the target KB at all.

Last, in parallel to this work \newcite{chen2017reading} evaluated an unstructured QA system against semantic parsing benchmarks. However, their focus was on examining the contributions of multi-task learning and distant supervision to training rather than to compare to state-of-the-art semantic parsers.

\section{Conclusion}
We propose in this paper to evaluate semantic parsing-based QA systems by comparing them to a web-based QA baseline. We evaluate such a QA system on \textsc{ComplexQuestions} and find that it obtains reasonable performance. We analyze performance and find that \textsc{Composition} and \textsc{Superlative} questions are challenging for a web-based QA system, while \textsc{Conjunction} and \textsc{N-ary} questions can often be handled by our QA model.

% \section{Related Work}

\paragraph{Reproducibility}
Code, data, annotations, and experiments
for this paper are available on the CodaLab platform
at \url{https://worksheets.codalab.org/worksheets/0x91d77db37e0a4bbbaeb37b8972f4784f/}.

\section*{Acknowledgments}
We thank Junwei Bao for providing us with the test predictions of his system.
We thank the anonymous reviewers
for their constructive feedback. This
work was partially supported by the Israel Science
Foundation, grant 942/16.

\bibliography{all}
\bibliographystyle{acl_natbib}

\end{document}